\documentclass[conference]{IEEEtran}
\usepackage{times}
\usepackage{todonotes}
\usepackage[numbers]{natbib}
\usepackage{multicol}
\usepackage{graphics,graphicx,caption,float,subcaption,booktabs,xcolor,multirow,array,color,ifthen,tabu,colortbl,dblfloatfix,url,xparse,mathtools,patchcmd,algorithm,algorithmic,amssymb,xspace,nicefrac,microtype,amsmath,amsfonts,bm,ragged2e,tikz,stackengine,etoolbox,xpatch,enumerate,xstring,setspace,tabularx,makecell,changepage,cuted,titlesec,wrapfig,tcolorbox, sidecap,comment, bbding, placeins}

\usepackage{gensymb,comment}
\usepackage[pagebackref=true,breaklinks=true,colorlinks=true,bookmarks=false,citecolor=blue]{hyperref}
\hypersetup{
colorlinks=true,
linkcolor=blue,
filecolor=magenta,      
citecolor=blue
}

\usepackage{siunitx}
\IEEEoverridecommandlockouts                              

\title{\LARGE \bf 
From Flow to One Step: \\
Real-Time Multi-Modal Trajectory Policies via Implicit Maximum Likelihood Estimation-based Distribution Distillation

}

\ifdefined\isanonymous
    \author{%
        Anonymous Authors
    \thanks{Affiliations withheld for double-blind review.
    }\\
    \thanks{
    }\\
    \thanks{
    }
    }
\else
    \author{
        Ju Dong$^{1,2,3*}$, 
        Liding Zhang$^{2}$,
        Lei Zhang$^{1,2,3*\dag}$,
        Yu Fu$^{2}$, 
        Kaixin Bai$^{1,3}$, \\ 
        Zoltán-Csaba Márton$^{3}$, 
        Zhenshan Bing$^{2}$, 
        Zhaopeng Chen$^{3}$, Alois Christian Knoll$^{2}$, Jianwei Zhang$^{1}$ 
        \thanks{\dag Corresponding author. lei.zhang-1@studium.uni-hamburg.de}
        \thanks{* The authors contributed equally to this work.}
        \thanks{$^{1}$TAMS (Technical Aspects of Multimodal Systems), Department of Informatics, University of Hamburg, Hamburg, Germany. 
        }
        \thanks{$^{2}$Technical University of Munich, Germany. 
        }
        \thanks{$^{3}$Agile Robots SE, Munich, Germany. 
        }
    
    }
\fi

\begin{document}
    \newcommand{\our} {LEAP Hand\xspace}
    \newcommand{\ourablation}{LEAP-C Hand\xspace}
    \newcommand{\ourshort}{LEAP\xspace}
    \def\@onedot{\ifx\@let@token.\else.\null\fi\xspace}
    \DeclareRobustCommand\onedot{\futurelet\@let@token\@onedot}
    \newcommand{\figref}[1]{Fig\onedot~\ref{#1}}
    \def\etal{\emph{et al}\onedot}
    \newcommand{\secref}[1]{Sec\onedot~\ref{#1}}
    \newcommand{\tabref}[1]{Tab\onedot~\ref{#1}}
    \newcommand\ananye[1]{\textcolor{red}{#1}}
    \makeatletter
    \let\@oldmaketitle\@maketitle
    \renewcommand{\@maketitle}{\@oldmaketitle
    \includegraphics[width=0.95\linewidth]{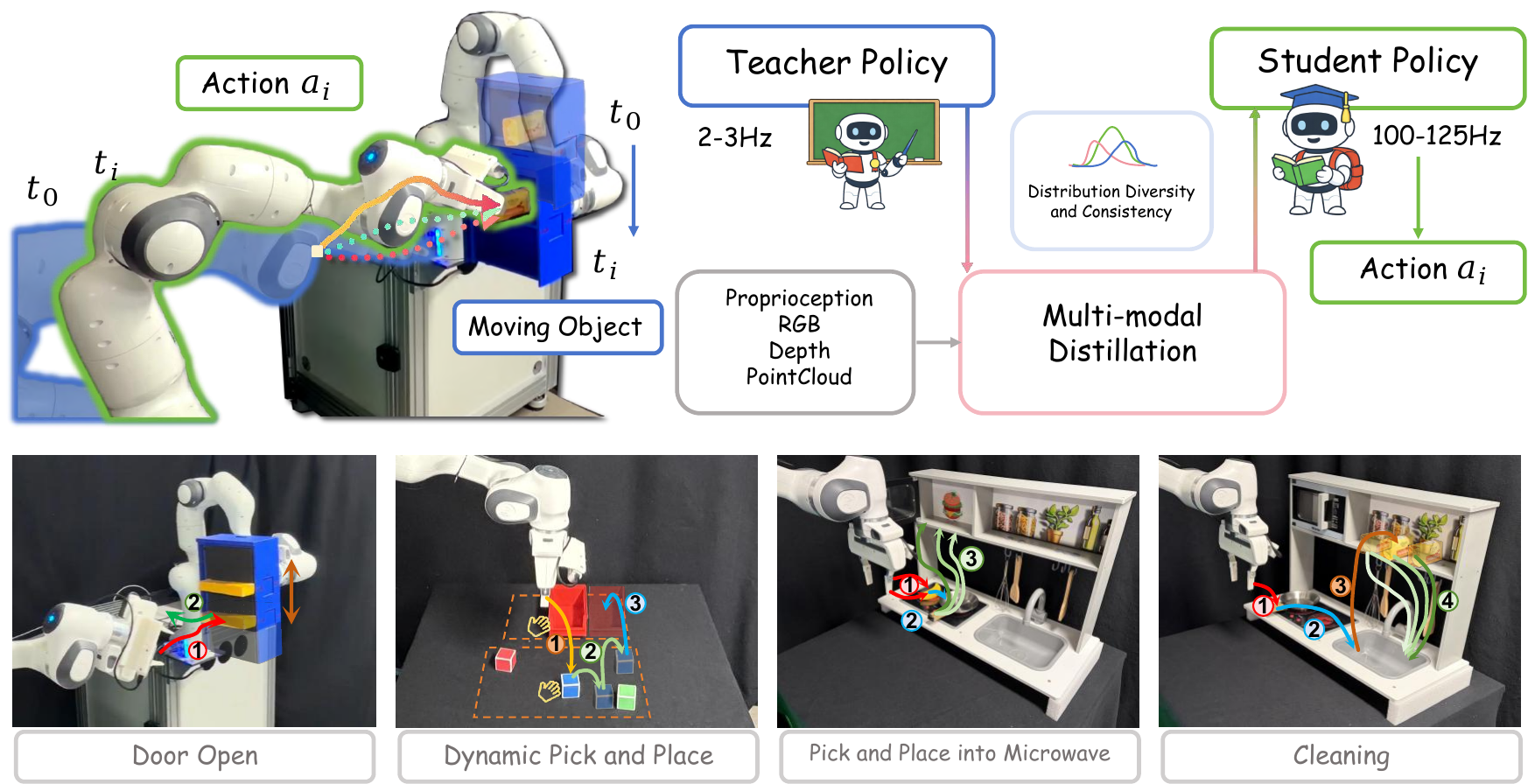}
      \centering
      \captionof{figure}{Overview of the distribution-level distillation framework and data diversity. \textit{Top:} Teacher–student distillation framework and student-generated multi-modal trajectory distributions at inference time. \textit{Bottom:} Example manipulation tasks demonstrating diverse trajectory data collected during training.
      }
      \label{fig:teaser}
      \vspace{-0.1in}
      \bigskip}
\makeatother
\maketitle
\thispagestyle{empty}
\pagestyle{empty}

\maketitle
\thispagestyle{empty}
\pagestyle{empty}
\setcounter{figure}{1}
\begin{abstract}
Generative policies based on diffusion and flow matching achieve strong performance in robotic manipulation by modeling multi-modal human demonstrations. However, their reliance on iterative Ordinary Differential Equation (ODE) integration introduces substantial latency, limiting high-frequency closed-loop control. Recent single-step acceleration methods alleviate this overhead but often exhibit distributional collapse, producing averaged trajectories that fail to execute coherent manipulation strategies. 
We propose a framework that distills a Conditional Flow Matching (CFM) expert into a fast single-step student via Implicit Maximum Likelihood Estimation (IMLE). A bi-directional Chamfer distance provides a set-level objective that promotes both mode coverage and fidelity, enabling preservation of the teacher’s multi-modal action distribution in a single forward pass. A unified perception encoder further integrates multi-view RGB, depth, point clouds, and proprioception into a geometry-aware representation. 
Extensive experiments demonstrate a strong balance between accuracy and speed. On RLBench, the student achieves 68.6\% success at 123.5 Hz, clearly outperforming dedicated one-step baselines such as Consistency Policy (16.3\%). In real-world deployment, it attains 70.0\% success at 125.0 Hz, providing a 43$\times$ speedup over the multi-step teacher (2.9 Hz) and enabling success on dynamic tasks where the teacher fails. The resulting high-frequency control supports real-time receding-horizon re-planning and improved robustness under dynamic disturbances. 
More details will be found at:  \url{https://sites.google.com/view/flow2one}.
\end{abstract}

\section{Introduction}
Learning robust and reactive manipulation policies from demonstrations remains a fundamental challenge in robotics. In dynamic environments, policies must process high-dimensional multimodal observations (e.g., RGB, depth, and point clouds) while handling the inherent stochasticity of human demonstrations, where multiple geometrically distinct trajectories can achieve the same goal. Traditional behavior cloning often averages conflicting demonstrations into physically implausible mean trajectories~\cite{gu2024bagail}. Consequently, generative models such as diffusion~\cite{chi2023diffusionpolicy,xu2025funcanon} and flow-matching~\cite{zhang2024flowpolicy} have gained prominence for modeling multi-modal action distributions.

Despite their expressiveness, diffusion- and flow-based policies suffer from significant computational overhead due to iterative denoising or ODE integration. In practice, inference speeds are often limited to 8–10 Hz in simulation and 2–3 Hz on physical robots~\cite{Ze2024DP3}, restricting high-frequency closed-loop control. Recent acceleration approaches, including naive single-step truncation and consistency distillation, attempt to remove iterative sampling but frequently exhibit mode collapse, where the policy averages over feasible strategies and fails to execute coherent manipulation behaviors.

To reconcile multi-modal expressiveness with real-time reactivity, we propose a generative framework centered on an IMLE-distilled single-step student policy. Policy distillation is formulated under an Implicit Maximum Likelihood Estimation (IMLE) objective using a sample-based, likelihood-free formulation. A bi-directional Chamfer distance serves as a set-level approximation, encouraging both mode coverage and mode fidelity while mitigating collapse~\cite{gx2025kl}.

To supervise the student, we construct an offline synthetic dataset using a Multimodal Conditional Flow Matching teacher. The teacher models expert behavior via a data-space predictive objective with flow-time-dependent masking, enabling accurate multi-step density modeling. We further introduce a unified perception architecture that fuses RGB, depth, point clouds, and proprioception into a geometry-aware representation.

Extensive simulation and real-world experiments validate our approach. 
In RLBench, the single-step student achieves 68.6\% success at 123.5 Hz, retaining approximately 93\% of the 50-step teacher’s performance (74.1\%) with a 14.3$\times$ speedup. 
In real-world deployment, the student achieves 70.0\% success at 125.0 Hz, providing a 43$\times$ speedup over the teacher (2.9 Hz). 
The resulting high-frequency control enables real-time receding-horizon re-planning and improves robustness to dynamic disturbances.

\textbf{Contributions:}
\begin{itemize}
    \item We propose a set-level IMLE-based distribution distillation framework that compresses a multi-step CFM expert into a single-step student policy for real-time multimodal control. A bi-directional Chamfer objective aligns action distributions at the set level, promoting both mode coverage and mode fidelity to mitigate mode collapse during single-step inference.

    \item We develop an integrated multimodal learning system that combines a Conditional Flow Matching teacher with a geometry-aware perception module, enabling stable policy training from heterogeneous sensory inputs.

    \item Extensive simulation and real-world experiments demonstrate real-time single-step inference (125 Hz) while achieving strong task performance, enabling robust high-frequency control for dynamic manipulation.
\end{itemize}

\section{Related Work}
\label{sec:related}
\subsection{Multimodal Perception for Robotic Manipulation}
Robotic manipulation requires visual representations that encode both semantic appearance and precise 3D geometry. RGB images provide dense texture cues, whereas depth maps and point clouds supply metric structure essential for spatial reasoning~\cite{shridhar2021cliportpathwaysroboticmanipulation,qi2017pointnetdeephierarchicalfeature}. Early imitation-learning systems typically fused these modalities via simple concatenation or independent encoders~\cite{zhao2023learning}, which limited their ability to establish cross-modal correspondences or mitigate modality-specific noise. As a result, RGB-based policies remain sensitive to lighting and occlusions, while depth and point-cloud observations often suffer from sparsity or missing regions. More advanced fusion strategies employ multi-view rendering or cross-attention to align 2D and 3D signals~\cite{dasari2024ditpi,dong2025m4diffusermultiviewdiffusionpolicy}. However, these approaches primarily emphasize feature aggregation rather than explicit interaction between modalities, and they generally lack mechanisms for adaptively weighting modalities when sensor reliability varies~\cite{chen2025acditadaptivecoordinationdiffusion}. Consequently, current multimodal perception pipelines still underutilize complementary geometric and semantic cues, which limits their effectiveness when conditioning high-capacity generative or policy models~\cite{jia2025pointmappolicystructuredpointcloud,he2025falcon}.

\subsection{Generative Models in Robotics}
Generative models have become a powerful paradigm for representing complex, multi-modal action distributions in robotic manipulation~\cite{zhang2025contactdexnet}. Diffusion-based policies~\cite{dasari2024ditpi,wang2024gendp3dsemanticfields,3d_diffuser_actor} generate trajectories through iterative refinement and naturally encode multiple plausible futures. While highly expressive, diffusion methods require tens to hundreds of denoising steps, resulting in low inference bandwidth and preventing deployment in real-time manipulation settings. Even when conditioned on multimodal sensory observations. Several acceleration techniques—including DDIM sampling~\cite{ze2024humanoid_manipulation}, higher-order solvers~\cite{lu2022dpmsolverfastodesolver,zhao2024dcsolverimprovingpredictorcorrectordiffusion}, and consistency distillation ~\cite{prasad2024consistency}—reduce the number of refinement steps but still rely on multiple evaluations of the generative dynamics and may compromise trajectory-level multi-modality.

A complementary line of work explores flow-based generative models~\cite{lipman2023flowmatchinggenerativemodeling,dai2025safeflowsaferobotmotion}, which learn continuous transport maps without requiring a diffusion noise schedule. Flow-based policies can represent multi-modal trajectory distributions, and recent variants show strong potential for long-horizon planning and multi-modal trajectory generation~\cite{fu2025moflowonestepflowmatching}. Nonetheless, flow-based inference requires solving an ODE, typically through multiple integration steps, incurring nontrivial runtime costs. Recent attempts to accelerate flow inference~\cite{yang2024consistencyflowmatchingdefining,zhang2024flowpolicy}—reduce solver steps but still rely on evaluating flow dynamics at multiple time points, without explicitly preserving diverse trajectory distributions.
A direction explores sampling-free generative policies based on Implicit Maximum Likelihood Estimation~\cite{rana2025imle}, which bypass both diffusion denoising and flow-based ODE integration by learning a direct mapping from noise to action trajectories. However, existing variants remain limited to low-dimensional behavior cloning and struggle in multi-modal manipulation, often exhibiting mode switching and lacking distribution-level matching needed for diverse trajectory prediction. Moreover, IMLE training relies on nearest-neighbor searches, which scale poorly with high-dimensional multimodal observations.

Moreover, existing diffusion- and flow-based controllers have been evaluated mostly under limited sensory conditioning and do not satisfy the high-frequency requirements of real-world manipulation.

\subsection{Distillation of Generative and Robotic Policies}

Distillation offers a promising way to accelerate generative models when full inference is too slow. Standard diffusion distillation compresses multi-step denoising into lightweight predictors. However, these methods commonly depend on regression objectives or KL divergence, which tend to average the distribution. This leads to mode collapse, where the student policy suppresses the natural diversity of the teacher. A similar issue arises in behavior cloning, where students often imitate averaged or single-mode behaviors rather than the full range of expert actions.

Recent approaches~\cite{prasad2024consistency,wang2024onestepdiffusionpolicyfast} attempt to bypass iterative sampling by learning auxiliary networks or enforcing consistency constraints. However, these methods introduce additional complexity and require evaluating the teacher multiple times. They also tend to become unstable for highly multi-modal distributions, where score fields are often ill-defined near mode boundaries. Furthermore, distillation for flow-based models remains limited; existing works typically assume unimodal outputs or require heavy computations, making them unsuitable for high-dimensional trajectory spaces.

To date, no prior work has effectively distilled multi-modal, long-horizon trajectory generators using rich sensory inputs. Bridging this gap is essential for real-time, robust manipulation.

\section{Problem Statement and Methods}\label{sec:method}

\subsection{Problem Statement}

We study real-time multimodal trajectory generation for robotic manipulation. 
At each control step $t$, the robot receives a high-dimensional observation 
$\mathbf{o}_t = \{\mathcal{I}_{rgb}, \mathcal{I}_{depth}, \mathcal{P}, \mathcal{S}\}$ 
consisting of multi-view RGB images, aligned depth maps, a 3D point cloud, and proprioceptive states. 

Our goal is to generate a set of diverse and temporally coherent future action trajectories 
$\{\boldsymbol{\tau}_k\}_{k=1}^K$, each with horizon $H$, conditioned on the current observation.

This problem presents three fundamental challenges:

(1) \textbf{Multimodal perception}, requiring robust fusion of heterogeneous semantic and geometric signals;

(2) \textbf{Multi-modal action generation}, requiring expressive distribution learning to capture diverse behaviors and prevent mode collapse;

(3) \textbf{Real-time inference}, demanding high-frequency closed-loop control, which precludes iterative sampling procedures.

While diffusion- and flow-based policies can model rich multimodal distributions, they rely on iterative denoising or ODE integration, typically limiting inference to $3$–$10$ Hz. 
Our objective is to retain their multimodal expressiveness while enabling instantaneous single-step inference suitable for real-time deployment.

\subsection{Method Overview}

Our key idea is to decouple expressive generative modeling from real-time control through distribution-level distillation.

We first train a powerful Conditional Flow Matching teacher that accurately models a multimodal trajectory distribution in data space. 
We then distill this distribution into a one-step student policy using a set-level Implicit Maximum Likelihood Estimation objective. 
The student achieves instantaneous inference while preserving the full multimodal diversity of the teacher.

The framework consists of three components:

\begin{enumerate}
    \item A Conditional Flow Matching teacher that learns a multimodal trajectory distribution;
    \item An IMLE-based one-step student policy that reconstructs this distribution without iterative sampling;
    \item A multimodal observation encoder that provides geometry-aware conditioning shared by both teacher and student.
\end{enumerate}

Unlike prior works that deploy generative models directly at inference time, we treat the teacher purely as an offline distribution oracle. 
The student is trained exclusively on teacher-generated trajectory sets, requiring neither score matching nor multi-time supervision. 
This design preserves distributional richness while collapsing inference to a single forward pass.

\begin{figure*}[t]
    \centering
    \includegraphics[width=0.9\linewidth]{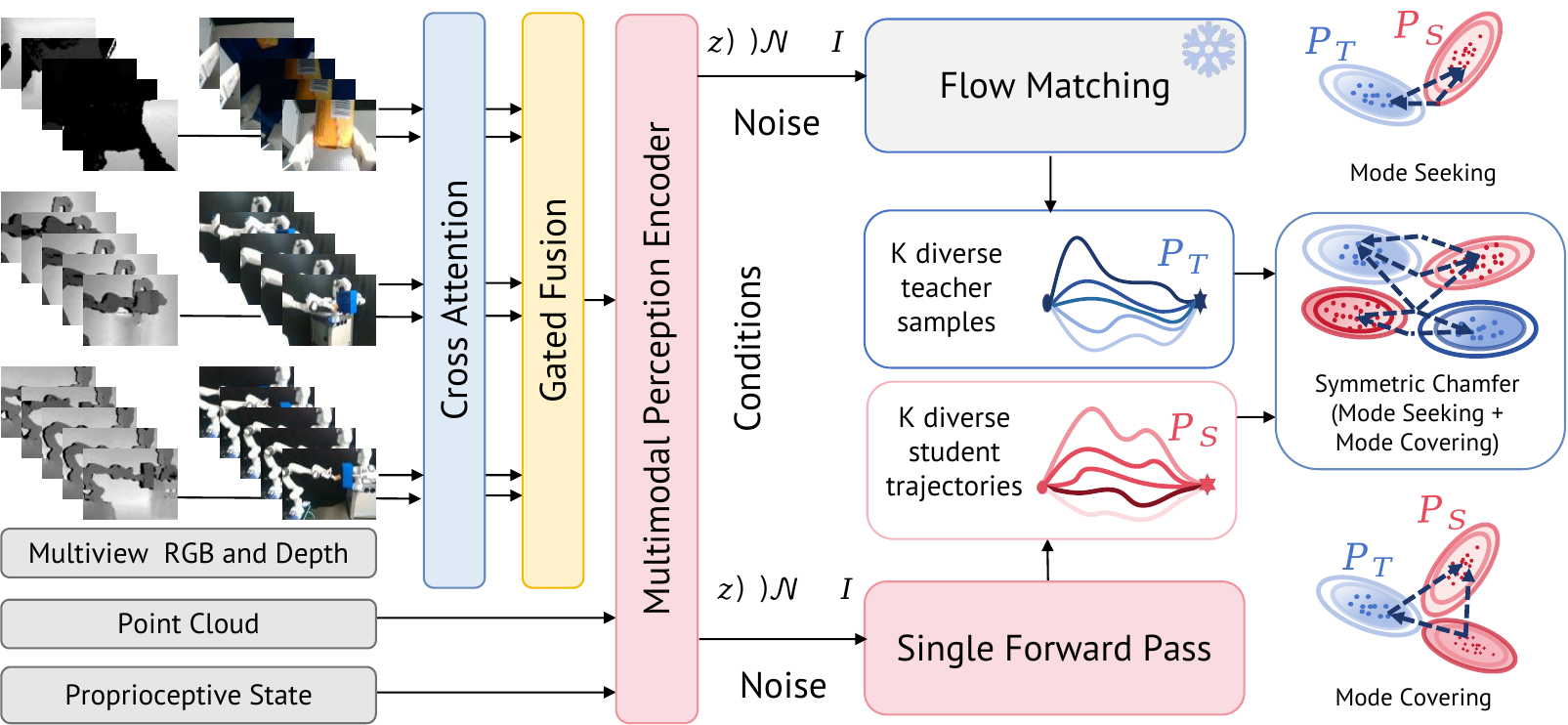}
    \caption{
    Detailed training pipeline. A unified multimodal encoder conditions both the ODE-based CFM teacher and the single-step student. The student is distilled from diverse teacher trajectories using set-level IMLE with a bi-directional Chamfer objective.
    }
    \label{fig:framework}
\end{figure*}

\subsection{IMLE-Based One-Step Policy Distillation}

To eliminate the computational bottleneck of flow-based sampling, 
we compress the teacher distribution into a single-step student policy 
$\pi_{\text{student}}$.

\subsubsection{One-Step Student Architecture.}

The student adopts an identical temporal 1D U-Net architecture to the teacher 
to strictly isolate improvements arising from the distillation algorithm rather than model capacity.

However, all time-conditioning modules (e.g., sinusoidal encodings and FiLM projections) are removed. 
The student directly maps a Gaussian noise vector 
$\mathbf{z} \sim \mathcal{N}(\mathbf{0}, \mathbf{I})$ 
and observation embedding $\mathbf{E}_{obs}$ to a complete trajectory:

\begin{equation}
\hat{\boldsymbol{\tau}} = \tau_\psi(\mathbf{E}_{obs}, \mathbf{z})
\in \mathbb{R}^{H \times D_{\text{action}}}.
\end{equation}

During deployment, students operated using a high-frequency control scheme: 
only the first $T_e$ steps are executed before re-planning with updated observations.

\subsubsection{Set-Level IMLE Distillation.}

Naive objectives such as MSE or KL divergence collapse multimodal outputs into mean trajectories. 
To prevent this, we formulate distillation as an Implicit Maximum Likelihood Estimation problem at the set level.

For each observation, the teacher provides a discrete set of $K$ multimodal trajectories:

\[
\mathcal{T}_{\text{teacher}} = \{\boldsymbol{\tau}_i^*\}_{i=1}^K.
\]

The student generates $K$ hypotheses:

\[
\{\hat{\boldsymbol{\tau}}_j\}_{j=1}^K.
\]

We minimize a symmetric Chamfer distance:

\begin{align}
\mathcal{L}_{\text{Chamfer}} &=
\frac{1}{K} \sum_{i=1}^K \min_j 
\left\| \boldsymbol{\tau}_i^* - \hat{\boldsymbol{\tau}}_j \right\|_2^2
\nonumber \\
&\quad +
\frac{1}{K} \sum_{j=1}^K \min_i 
\left\| \boldsymbol{\tau}_i^* - \hat{\boldsymbol{\tau}}_j \right\|_2^2.
\end{align}

The first term ensures \textbf{mode covering}, guaranteeing that each teacher trajectory is matched by at least one student hypothesis. 
The second term enforces \textbf{mode seeking}, preventing spurious generations outside the teacher's behavior manifold.

By minimizing this objective, the student reconstructs the geometric and statistical diversity of the teacher distribution while requiring only a single forward pass at inference time.

\subsection{Conditional Flow Matching Teacher}

To construct high-quality multimodal supervision for distillation, 
we train a Conditional Flow Matching teacher to model the trajectory distribution in data space.

\subsubsection{Flow Matching Objective.}

Let $\boldsymbol{\tau}_1$ denote a ground-truth trajectory sampled from expert data 
$q(\boldsymbol{\tau})$, and let 
$\boldsymbol{\tau}_0 \sim \mathcal{N}(\mathbf{0}, \mathbf{I})$ 
be a noise trajectory of the same dimension.

CFM defines a linear probability path:

\begin{equation}
\boldsymbol{\tau}_t = (1 - t)\boldsymbol{\tau}_0 + t\boldsymbol{\tau}_1,
\quad t \in [0,1].
\end{equation}

The teacher network $D_\theta$ predicts the target trajectory directly in data space:

\begin{equation}
\mathcal{L}_{\text{CFM}} =
\mathbb{E}_{t, \boldsymbol{\tau}_0, \boldsymbol{\tau}_1}
\left\|
D_\theta(\boldsymbol{\tau}_t, t, \mathbf{E}_{obs})
- \boldsymbol{\tau}_1
\right\|_2^2.
\end{equation}

\subsubsection{Time Scheduling and Anti-Shortcut Regularization.}

We sample $t$ from a logit-normal distribution to allocate more capacity to early, noisy stages. 
To prevent trivial identity mappings as $t \to 1$, 
we introduce a time-dependent masking mechanism that stochastically suppresses the noise embedding, 
forcing the model to rely on the observation condition.

\subsubsection{Teacher Trajectory Sampling.}

After training, we generate $K$ multimodal trajectories per observation by solving the corresponding ODE defined by $D_\theta$. 
This produces an explicit discrete trajectory set:

\[
\mathcal{T}_{\text{teacher}} = 
\{\boldsymbol{\tau}_1^{*(1)}, \dots, \boldsymbol{\tau}_1^{*(K)}\}.
\]

These sets form the supervision targets for IMLE distillation. 
The teacher is used exclusively offline and never deployed during real-time control.

\subsection{Multimodal Observation Encoding}

Both teacher and student are conditioned on a shared multimodal embedding 
$\mathbf{E}_{obs}$. 
We design a geometry-aware encoder to robustly fuse semantic and structural signals.

\subsubsection{Visual Encoding.}

RGB and depth images are processed by dual ResNet-18 backbones. 
Multi-view features are projected into a shared latent space, 
yielding embeddings $\mathbf{f}_{rgb}$ and $\mathbf{f}_{depth}$.

\subsubsection{Bi-Directional Cross-Modal Attention.}

We introduce symmetric cross-attention:

\begin{align}
\mathbf{f}'_{rgb} &= \mathbf{f}_{rgb} 
+ \text{Attn}(\mathbf{f}_{rgb}, \mathbf{f}_{depth}, \mathbf{f}_{depth}), \\
\mathbf{f}'_{depth} &= \mathbf{f}_{depth} 
+ \text{Attn}(\mathbf{f}_{depth}, \mathbf{f}_{rgb}, \mathbf{f}_{rgb}),
\end{align}

establishing dense semantic-geometric correspondences.

\subsubsection{Adaptive Fusion and Global Embedding.}

A gating network predicts modality weights 
$\boldsymbol{\alpha} = [\alpha_{rgb}, \alpha_{depth}]$ 
for noise-robust fusion.

Point clouds are encoded via PointNet to obtain $\mathbf{h}_{pcd}$, 
and proprioception is embedded using an MLP into $\mathbf{h}_{state}$.

The final observation embedding is

\begin{equation}
\mathbf{E}_{obs} = 
[\mathbf{h}_{vis}, \mathbf{h}_{pcd}, \mathbf{h}_{state}],
\end{equation}

which conditions both teacher and student policies.

\section{Experiment}
\label{sec:experiment}
To evaluate our proposed framework, we conduct extensive experiments in both simulation and real-world scenarios. Our evaluation aims to address three key questions: (i) Does the multimodal CFM teacher provide a robust expert trajectory distribution for complex manipulation? (ii) Can the IMLE-distilled student policy maintain high success rates while enabling high-frequency, real-time inference? (iii) How does our one-step method compare against original multi-step generative experts, their naive one-step variants, and other dedicated distillation baselines?

\subsection{Simulation Experiment}
We conduct simulation experiments on the RLBench benchmark~\cite{james2019rlbench}. Following the setup in PointFlowMatch~\cite{chisari2024learning}, we adopt an identical set of eight representative manipulation tasks.

\subsubsection{Implementation Details} 
\textbf{Baselines.} We compare our method against three categories of generative policies. For the iterative models, we evaluate both their original multi-step versions and their naive 1-step inference variants: 
(i) \textbf{Standard diffusion-based policies}: Diffusion Policy~\cite{chi2023diffusionpolicy} and Diffusion Policy 3D (DP3)~\cite{Ze2024DP3}; 
(ii) \textbf{Flow-matching-based approaches}: PointFlowMatch~\cite{chisari2024learning} and FlowPolicy~\cite{zhang2024flowpolicy}; and 
(iii) \textbf{Dedicated one-step generative baselines}: Consistency Policy~\cite{prasad2024consistency} (via distillation) and \textbf{IMLE Policy}~\cite{rana2025imle} (direct IMLE on demonstrations).

\textbf{Data Collection and Training.} For each task, 100 expert demonstrations are collected using the RLBench motion planner. The CFM teacher is trained for 2,000 epochs. Using the frozen teacher, we generate K=16 trajectories of horizon H=32 per observation to construct the distilled dataset. The single-step student is trained on this dataset for 1,500 epochs.

\textbf{Evaluation.} All models are evaluated on 300 episodes with randomized initial states. Results are averaged over three independent runs, each consisting of 300 evaluation episodes.

\subsubsection{Results Analysis}
\begin{figure} \centering \includegraphics[width=0.95\linewidth]{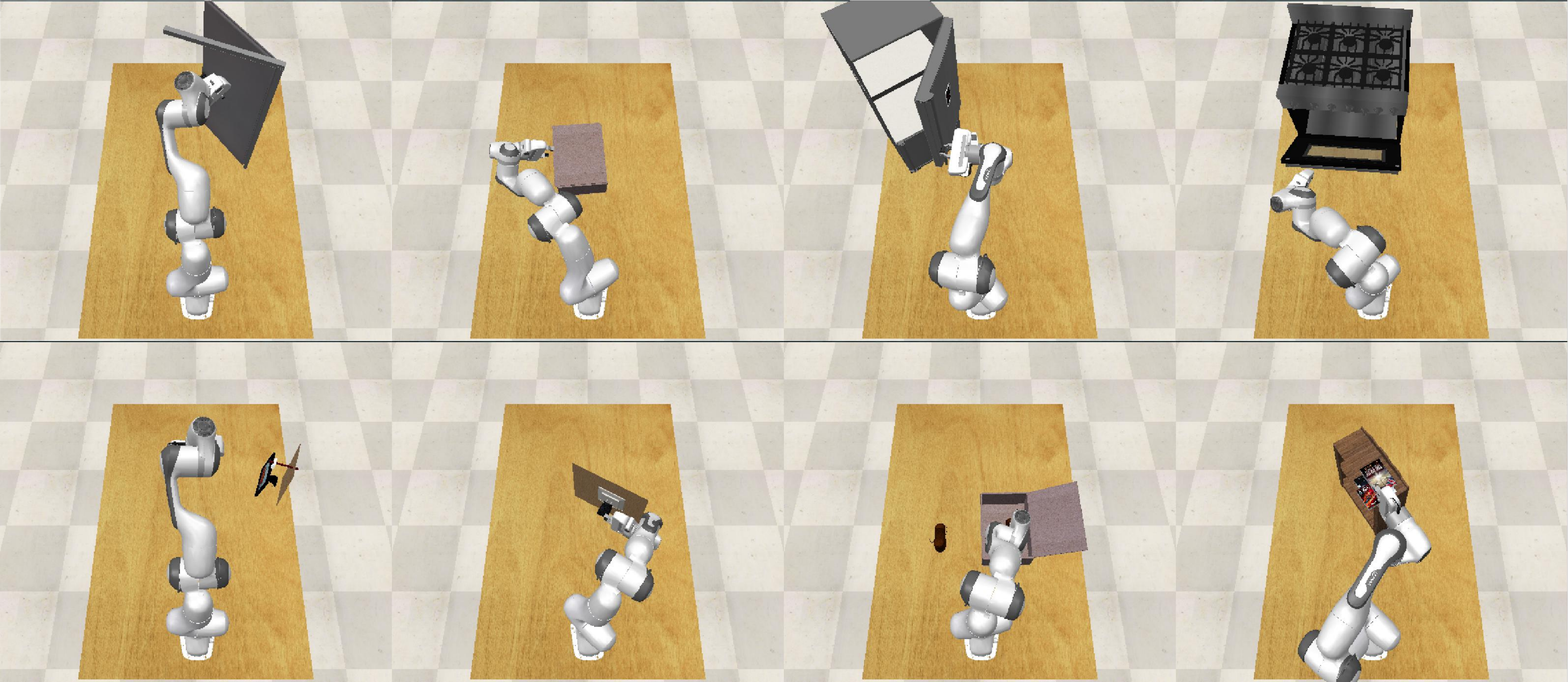} \caption{Qualitative results on RLBench tasks.
Tasks are ordered from left to right and top to bottom: close door, open box, open fridge, open oven, take frame off hanger, unplug charger, take shoes out of box, and put books on bookshelf.}\label{fig:RLBench} \end{figure}

\textbf{Qualitative Results.}
Fig.~\ref{fig:RLBench} shows successful executions of the distilled student policy across eight RLBench tasks, demonstrating robust performance under randomized object initializations.
\begin{table*}[t]
\centering
\caption{Quantitative comparison of success rate and inference speed across simulated manipulation tasks.}
\label{tab:sim}
\resizebox{0.9\textwidth}{!}{%
\begin{tabular}{l|cccccccc|cc}
\toprule
Method 
& unplug charger & close door & open box & open fridge 
& frame hanger & open oven & books shelf & shoes box 
& Avg SR $\uparrow$ & Avg IS $\uparrow$ \\
\midrule
Diffusion Policy (50-step)
& 38.0$\pm$3.6 & 19.3$\pm$2.5 & 75.7$\pm$4.2 & 0.0$\pm$0.0 & 16.0$\pm$2.6 & 0.0$\pm$0.0 & 0.3$\pm$0.6 & 0.0$\pm$0.0 & 18.7$\pm$2.3 & 8.63 \\
Diffusion Policy (1-step)
& 5.1$\pm$1.0 & 3.1$\pm$0.7 & 2.1$\pm$0.8 & 0.0$\pm$0.0 & 4.2$\pm$1.2 & 0.0$\pm$0.0 & 0.0$\pm$0.0 & 0.0$\pm$0.0 & 1.8$\pm$0.5 & 95.07 \\
\addlinespace
DP3 (50-step)
& 33.3$\pm$4.7 & 76.0$\pm$1.7 & 98.3$\pm$1.5 & 4.3$\pm$2.1 & 12.3$\pm$2.5 & 0.3$\pm$0.6 & 3.7$\pm$0.6 & 0.0$\pm$0.0 & 28.5$\pm$2.2 & 8.02 \\
DP3 (1-step)
& 13.2$\pm$5.2 & 25.2$\pm$4.0 & 35.1$\pm$5.9 & 3.3$\pm$1.0 & 4.6$\pm$3.0 & 0.0$\pm$0.0 & 1.4$\pm$1.7 & 0.0$\pm$0.0 & 10.4$\pm$2.5 & 220.06 \\
\addlinespace
PointFlowMatch (50-step)
& 83.6$\pm$3.3 & 68.3$\pm$6.6 & 99.4$\pm$0.7 & 31.9$\pm$2.9 & 38.6$\pm$2.7 & 75.9$\pm$4.0 & 68.8$\pm$5.8 & 76.0$\pm$3.5 & 67.8$\pm$4.1 & 9.12 \\
PointFlowMatch (1-step)
& 46.3$\pm$5.2 & 37.4$\pm$5.3 & 54.1$\pm$3.2 & 17.8$\pm$5.0 & 21.6$\pm$4.5 & 49.8$\pm$4.7 & 59.2$\pm$5.4 & 16.1$\pm$3.4 & 37.8$\pm$4.3 & 207.12 \\
\addlinespace
FlowPolicy (50-step)
& 85.7$\pm$2.9 & 53.3$\pm$5.4 & 97.1$\pm$1.2 & 29.7$\pm$4.8 & 37.3$\pm$4.2 & 78.4$\pm$3.7 & 70.2$\pm$5.1 & 62.8$\pm$4.6 & 64.3$\pm$3.5 & 9.97 \\
FlowPolicy (1-step)
& 52.2$\pm$2.7 & 35.3$\pm$6.0 & 44.1$\pm$5.4 & 20.2$\pm$5.1 & 24.7$\pm$4.6 & 47.8$\pm$5.7 & 46.8$\pm$5.3 & 12.9$\pm$5.8 & 35.5$\pm$4.8 & 216.43 \\
\addlinespace
IMLE Policy (1-step)
& 41.9$\pm$3.5 & 22.1$\pm$3.5 & 76.2$\pm$2.4 & 1.1$\pm$0.8 & 13.8$\pm$2.3 & 0.0$\pm$0.0 & 0.6$\pm$0.5 & 0.3$\pm$0.3 & 19.5$\pm$1.7 & 99.56 \\
Consistency Policy (1-step)
& 30.2$\pm$3.2 & 16.8$\pm$3.8 & 73.8$\pm$2.0 & 0.4$\pm$0.5 & 7.9$\pm$2.2 & 0.0$\pm$0.0 & 1.1$\pm$0.8 & 0.0$\pm$0.0 & 16.3$\pm$1.5 & 100.47 \\
\midrule
CFM-Teacher (50-step)
& 89.8$\pm$1.4 & 79.8$\pm$2.4 & 99.1$\pm$0.5 & 43.8$\pm$3.5 & 40.2$\pm$4.2 & 81.8$\pm$2.5 & 78.2$\pm$2.8 & 80.2$\pm$2.5 & 74.1$\pm$2.5 & 8.60 \\
\rowcolor{gray!10}CFM-Student (1-step)
& 84.1$\pm$1.5 & 73.8$\pm$2.7 & 97.9$\pm$0.5 & 35.2$\pm$3.7 & 36.9$\pm$3.5 & 76.2$\pm$2.8 & 69.8$\pm$3.7 & 75.1$\pm$3.0 & 68.6$\pm$2.7 & 123.5 \\
\bottomrule
\end{tabular}%
}
\end{table*}

\textbf{Quantitative Results.}
Table~\ref{tab:sim} reports the average success rate (Avg SR) and inference speed (Avg IS) across eight simulated manipulation tasks.
The results demonstrate the effectiveness of the proposed CFM teacher and IMLE-based distillation framework.

\textbf{Superiority of Single-Step Generation.}
Table~\ref{tab:sim} shows that diffusion and flow-based methods degrade substantially when their iterative sampling is reduced to a single step.
Diffusion Policy drops from 18.7\% (50-step) to 1.8\% (1-step).
PointFlowMatch and FlowPolicy decrease from 67.8\% to 37.8\%, and from 64.3\% to 35.5\%, respectively.
Consistency Policy achieves 16.3\% SR, indicating difficulty in modeling multi-modal action distributions.
In contrast, the IMLE-distilled CFM-Student (1-step) achieves 68.6\% Avg SR, outperforming all other 1-step baselines and remaining close to the 50-step CFM-Teacher (74.1\%).
This indicates that set-level distillation preserves most of the teacher’s capability under single-step inference.

\textbf{Balance Between Precision and Speed.}
The CFM-Student runs at 123.5 Hz, providing a 14.3× speedup over the CFM-Teacher (8.60 Hz).
Although some 1-step baselines achieve higher raw frequencies, their success rates remain substantially lower, reflecting difficulty in preserving multi-modal action structure under single-step inference.
The Student maintains high task performance while exceeding 100 Hz, which is commonly considered suitable for high-frequency closed-loop control.

\textbf{Task-Level Robustness.}
Across tasks, the Student consistently preserves multi-modal trajectory modeling.
On simpler tasks such as \textit{open box}, it achieves 97.9\%, close to the Teacher (99.1\%).
On more challenging long-horizon tasks such as \textit{open oven} and \textit{shoes box}, naive 1-step variants fail or degrade significantly (e.g., 0.0\% for Diffusion Policy), whereas the Student achieves 76.2\% and 75.1\%, respectively.
These results suggest that the bi-directional Chamfer objective mitigates distributional collapse under single-step inference.

\section{Ablation Studies}
\label{sec:ablation}
To investigate the contribution of individual components in our framework, we conduct ablation studies focusing on three key aspects: (1) the necessity of the trimodal sensor fusion strategy, (2) the number of distilled trajectory candidates $K$, and (3) the prediction horizon $T$. Results are averaged over the most challenging tasks and reported in Table~\ref{tab:ablation_unified}.

\begin{table}[t]
\centering
\footnotesize
\setlength{\tabcolsep}{6pt}
\caption{\textbf{Ablation Studies.} Impact of modality, distillation candidates $K$, and horizon $T$.}
\label{tab:ablation_unified}
\resizebox{0.25\textwidth}{!}{%
\begin{tabular}{lc}
\toprule
\textbf{Variant} & \textbf{SR (\%)} \\
\midrule
\multicolumn{2}{l}{\textit{(a) Modality}} \\
\midrule
RGB + Depth (Concat) & 30.4 \\
RGB + Depth (Attention) & 39.6 \\
RGB + Depth (Gated Attn.) & 66.2 \\
PCD + RGB & 60.7 \\
PCD + Depth & 59.8 \\
\rowcolor{gray!10}\textbf{Our Fusion} & \textbf{74.1} \\

\midrule
\multicolumn{2}{l}{\textit{(b) Distillation $K$}} \\
\midrule
$K = 1$ & 47.3 \\
$K = 4$ & 58.9 \\
$K = 10$ & 68.1 \\
\rowcolor{gray!10}\textbf{$K = 16$} & \textbf{74.1} \\

\midrule
\multicolumn{2}{l}{\textit{(c) Horizon $T$}} \\
\midrule
$T = 1$ & 30.7 \\
$T = 8$ & 64.8 \\
$T = 16$ & 70.9 \\
\rowcolor{gray!10}\textbf{$T = 32$} & \textbf{74.1} \\
\bottomrule
\end{tabular}
}
\end{table}

\textbf{Impact of Multi-Modal Fusion.}
Table~\ref{tab:ablation_unified}(a) shows that incorporating point cloud information consistently improves performance over purely 2D modalities.
The full three-modal fusion achieved the strongest results, demonstrating that geometric and visual cues make complementary contributions.

\textbf{Effect of Distillation Candidates $K$.}
As shown in Table~\ref{tab:ablation_unified}(b), performance improves monotonically with increasing $K$.
Larger candidate sets better capture the teacher’s multi-modal action distribution, leading to more effective distillation.

\textbf{Importance of Prediction Horizon $T$.}
Table~\ref{tab:ablation_unified}(c) demonstrates that longer prediction horizons yield consistently better performance.
Extending $T$ provides richer temporal context, enabling more coherent and robust manipulation behaviors.

\subsection{Real-World Experiments} 
\begin{figure}[t]
    \centering
    \includegraphics[width=0.32\linewidth]{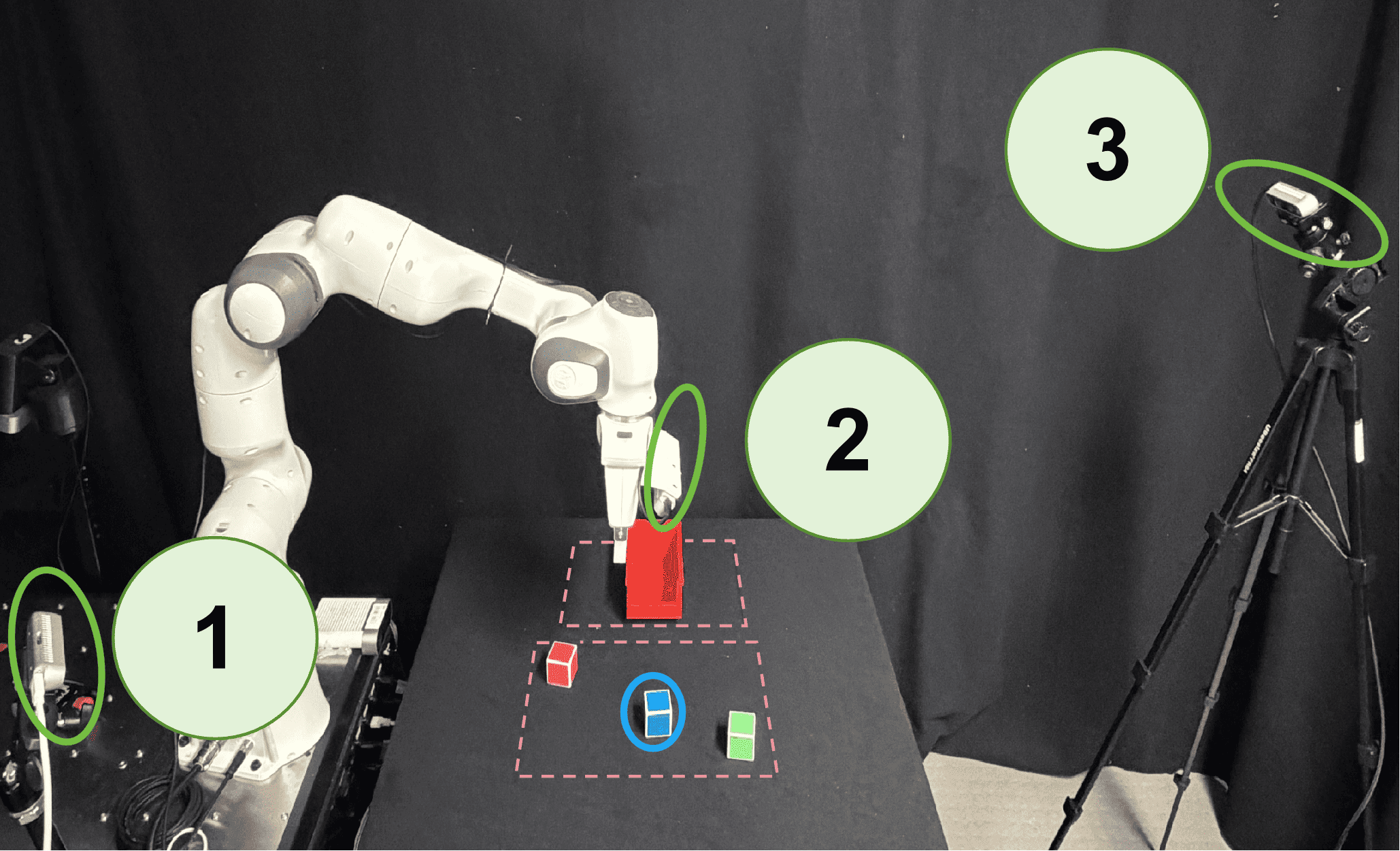}
    \hfill
    \includegraphics[width=0.32\linewidth]{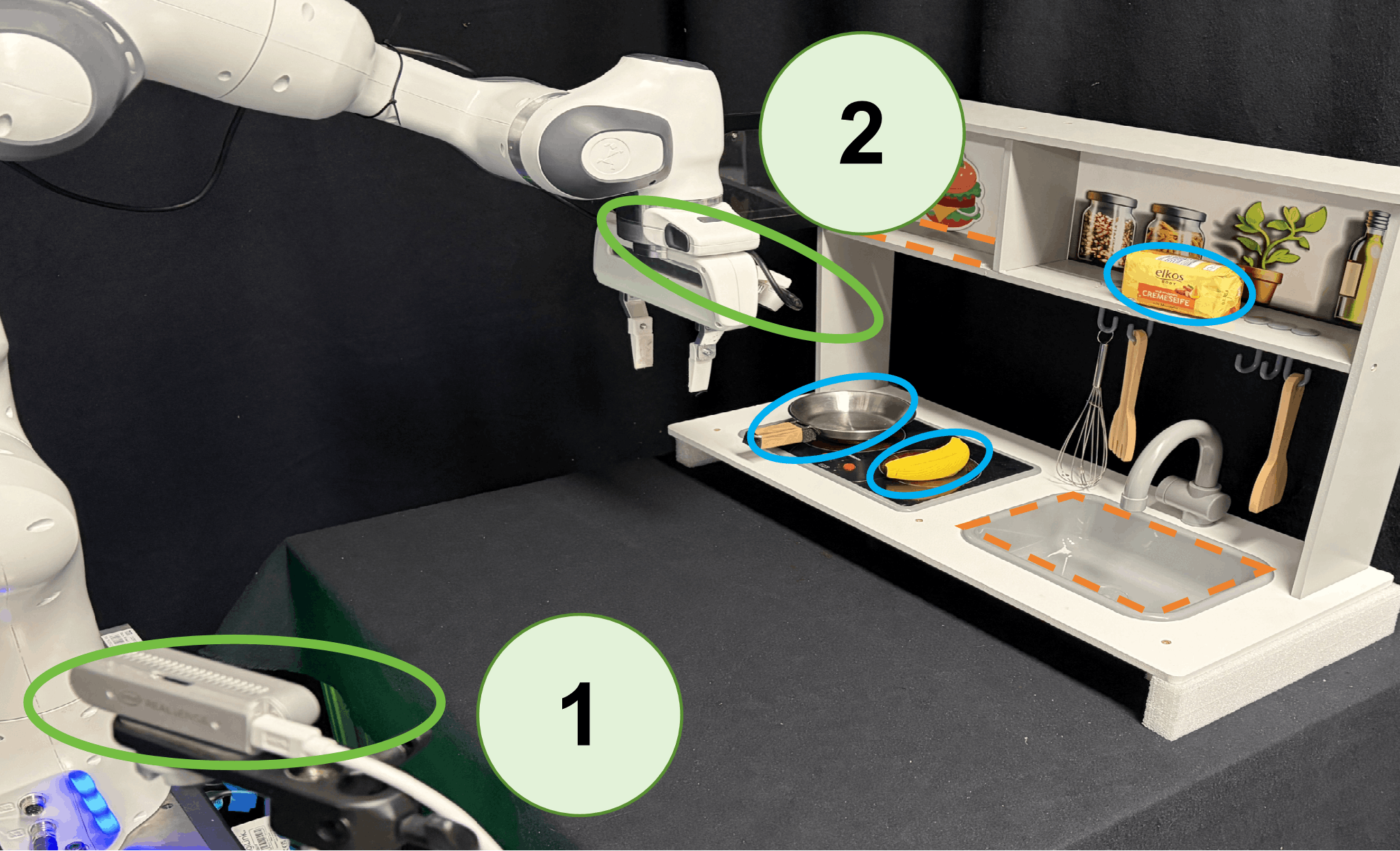}
    \hfill
    \includegraphics[width=0.32\linewidth]{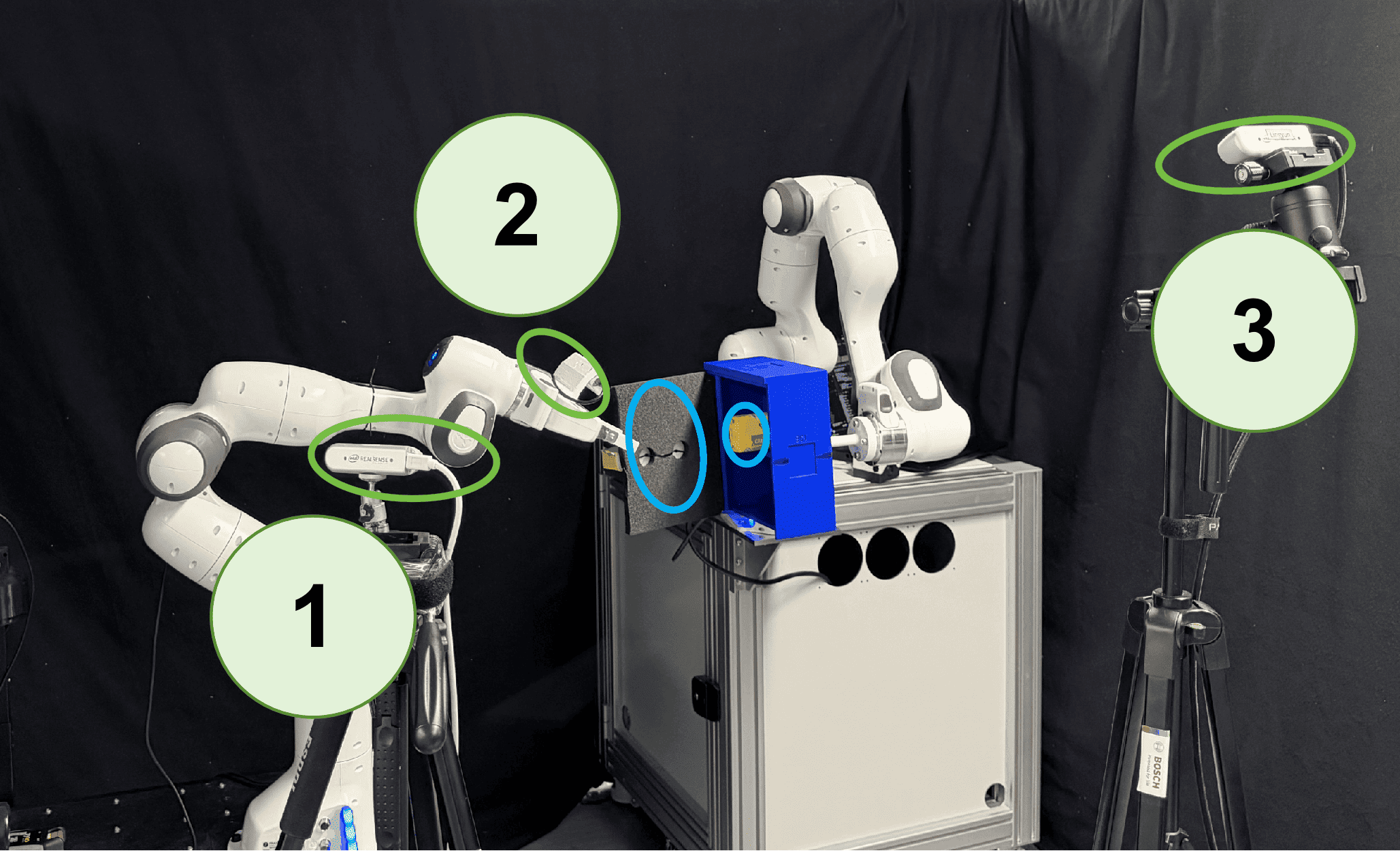}
    \caption{Three real-world setups. Cameras 1--3 correspond to the back, wrist, and external cameras, respectively.}
    \label{fig:real_setup}
\end{figure}

We design five real-world manipulation tasks covering dynamic perturbations, long-horizon multi-stage execution, and moving-object interaction.
The tasks are:
\begin{enumerate} \item \textbf{Dynamic Cube Stowing:} Grasping and placing under continuous human-induced perturbations.
\item \textbf{Kitchen Microwave Loading:} 
Long-horizon sequential manipulation including object transfer and container placement.
\item \textbf{Kitchen Cleanup:} 
Sequential rearrangement of multiple objects with varying spatial configurations.
\item \textbf{Dynamic Cabinet Opening:} 
Grasping and pulling a continuously moving cabinet door.
\item \textbf{Dynamic Grasping:} 
Reaching into a moving cabinet to retrieve a target object.

\end{enumerate}
Figure~\ref{fig:real_setup} illustrates the experimental setup. From left to right: Dynamic Cube Stowing, Kitchen Tasks, and Cabinet Tasks.

\subsubsection{Implementation Details}

\textbf{Hardware Setup.} Experiments are conducted on a 7-DoF Franka Emika Panda robot with a parallel gripper. Perception uses one wrist-mounted RGB-D camera and multiple external RGB-D cameras. Point clouds are cropped to the workspace and voxel-downsampled to 4,096 points.

\textbf{Data Collection.} For each task, we captured 300 remote operation demonstration trajectories using a 3D SpaceMouse under varying initial object configurations and control strategy settings. Data for the same task encompassed multiple feasible execution paths, ensuring diversity in spatial distribution and motion patterns for the teaching data.

\textbf{Training.} The multimodal CFM teacher is trained on demonstration data. Multi-modal trajectories are generated via ODE-based sampling to form a distilled supervision set. A single-step student policy is then optimized using IMLE with a bi-directional Chamfer loss. All architectures and hyperparameters follow those used in the simulation setup.

\textbf{Deployment.} Inference runs on an RTX 3070 GPU. Actions are sent to a real-time control PC via Ethernet. For dynamic cabinet tasks, an additional arm induces independent perturbations.

\textbf{Evaluation Protocol.} Each task is evaluated over 30 real-world trials with randomized initial conditions. We report success rate and inference speed (Hz).
\subsubsection{Results Analysis}
\begin{table*}[t]
\centering
\caption{
Quantitative results on real-robot manipulation tasks.
{\footnotesize \textit{($^*$ indicates tasks that cannot be completed due to the latency of multi-step inference.)}}
}
\label{tab:comparison}
\resizebox{0.95\textwidth}{!}{
\begin{tabular}{l|cc|ccc|cc}
\toprule
 & \multicolumn{2}{c|}{\textbf{Static Tasks}} 
 & \multicolumn{3}{c|}{\textbf{Dynamic Tasks}}
 &  &  \\
\textbf{Method} 
 & Microwave Loading 
 & Kitchen Cleanup 
 & Cube Stowing
 & Cabinet Opening 
 & Grasping 
 & \textbf{Avg SR $\uparrow$} 
 & \textbf{Avg IS $\uparrow$} \\
\midrule

PointFlowMatch (1-step)
& 6.7 & 6.7 
& 3.3 & 0.0 & 0.0 
& 3.3
& 70.0 \\

PointFlowMatch (50-step)
& 86.7 & 90.0
& 66.7 & 0.0$^*$ & 0.0$^*$
& 48.7
& 3.6 \\

\midrule

CFM-Teacher (50-step)
& 93.3 & 96.7
& 80.0 & 0.0$^*$ & 0.0$^*$
& 54.0
& 2.9 \\

\rowcolor{gray!10}
CFM-Student (1-step)
& 83.3 & 86.7
& 63.3 & 50.0 & 66.7
& 70.0
& 125.0 \\

\bottomrule
\end{tabular}
}
\end{table*}

\textbf{Overall Performance and Real-Time Capability.}
As shown in Table~\ref{tab:comparison}, the distilled Student achieves a strong balance between task performance and inference speed in real-world settings.
With an average inference speed of \textbf{125.0 Hz}, it is approximately \textbf{43$\times$ faster} than the CFM-Teacher (2.9 Hz) and about \textbf{35$\times$ faster} than the 50-step PointFlowMatch baseline (3.6 Hz).

\textbf{Comparison with Baselines.}
The \textit{PointFlowMatch (1-step)} baseline attains a relatively high inference speed (70.0 Hz) but collapses to a near-zero average success rate (\textbf{3.3\%}).
In contrast, the distilled \textbf{Student} achieves a robust \textbf{70.0\%} average success rate while maintaining single-step inference.

\textbf{Teacher Performance and Efficiency Trade-off.}
The iterative \textbf{CFM-Teacher (50-step)} achieves strong performance on multiple tasks, reaching \textbf{93.3\%} and \textbf{96.7\%} success rates on \textit{Microwave Loading} and \textit{Kitchen Cleanup}, respectively, and \textbf{80.0\%} on \textit{Cube Stowing}.
However, its inference frequency is limited to 2.9 Hz due to the multi-step sampling process.
In contrast, the distilled \textbf{Student} achieves \textbf{70.0\%} average success rate at \textbf{125.0 Hz}, demonstrating a significantly better balance between task performance and real-time control capability.

\textbf{Robustness in Dynamic Scenarios.}
Dynamic manipulation tasks highlight the advantage of fast single-step policies.
While both the 50-step PointFlowMatch baseline and the CFM-Teacher fail to complete \textit{Cabinet Opening} and \textit{Grasping}, the \textbf{Student} achieves success rates of \textbf{50.0\%} and \textbf{66.7\%}, respectively.
These results indicate that the distilled policy can effectively react to environmental perturbations and moving objects, which is difficult for slow iterative inference pipelines.
On \textit{Cube Stowing}, the Student achieves \textbf{63.3\%}, reflecting the increased difficulty under dynamic perturbations.

\begin{figure}[t]
\centering
\includegraphics[width=0.78\linewidth]{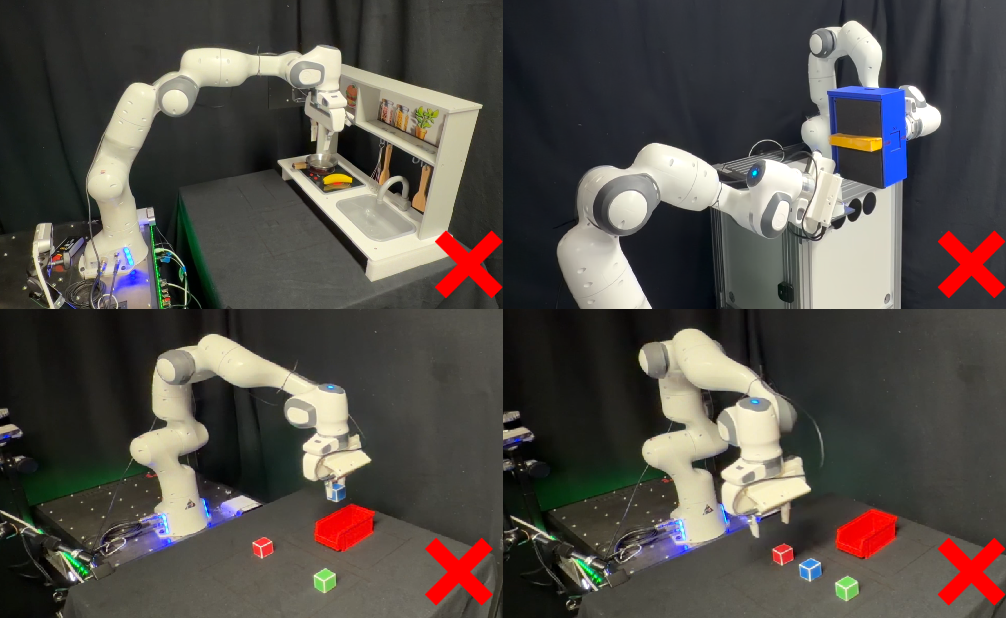}
\caption{Mode-Collapse Failures of the 1-Step Baseline.}
\label{fig:failure}
\end{figure}

\textbf{Failure Analysis.}
To analyze the failure of naive single-step acceleration, we examine the 1-step PointFlowMatch baseline on real robots.
Across dynamic tasks, we observe a consistent pattern: the policy reaches task-relevant regions but stalls at critical commitment points (Fig.~\ref{fig:failure}).
In Grasping, the gripper approaches the object yet fails to establish a stable grasp.
In Cube Stowing, the cube is lifted but not inserted into the target region.
In the Kitchen task, the robot oscillates between candidate objects without completing a pick-and-place cycle.

These behaviors are not characteristic of perception failure or control latency: the robot correctly localizes target objects and reaches feasible intermediate states without divergence. 
Rather, the pattern aligns with distributional averaging, where single-step truncation collapses the multi-modal action distribution and produces trajectories that fail to commit to a coherent manipulation plan.

Notably, such behaviors are largely absent in our IMLE-distilled \textbf{Student}, which successfully executes the same tasks (Table~\ref{tab:comparison}).

\begin{table}[t]
\centering
\caption{Failure mode breakdown over failed trials (count and percentage).}
\label{tab:failure_breakdown}
\small
\setlength{\tabcolsep}{6pt}
\resizebox{0.48\textwidth}{!}{%
\begin{tabular}{lcc}
\toprule
Failure Type & PointFlowMatch (1-step) & CFM-Student \\
\midrule
\rowcolor{gray!10}Mode collapse & 109 (75.1\%) & 3 (6.7\%) \\
Wrong target & 18 (12.4\%) & 12 (26.7\%) \\
Collision & 12 (8.3\%) & 11 (24.4\%) \\
Unstable grasp & 6 (4.1\%) & 19 (42.2\%) \\
\midrule
\textbf{Total failed trials} & \textbf{145} & \textbf{45} \\
\bottomrule
\end{tabular}
}
\end{table}

Table~\ref{tab:failure_breakdown} reports the failure mode distribution over failed trials.
\textbf{75.1\%} of failures from the naive baseline exhibit mode-collapse behaviors, whereas Student failures are dominated by collisions and contact instability.
This contrast suggests that naive single-step truncation primarily damages the multi-modal action distribution rather than low-level execution.

\section{Conclusion and Future Work}
\label{sec:conclusion}

In this paper, we present a novel framework for real-time multi-modal trajectory generation in robotic manipulation. By modeling the expert behavior distribution via a Conditional Flow Matching teacher, the framework captures rich multi-modal action distributions conditioned on a unified perceptual embedding. To eliminate the inherent latency of ODE-based flow sampling, we further introduce a sample-only, likelihood-free distillation approach based on Implicit Maximum Likelihood Estimation with a set-level Chamfer matching objective.

Crucially, by maintaining strict architectural consistency between the multi-step teacher and the single-step student, we rigorously demonstrated that our distillation algorithm effectively compresses the generative process without sacrificing capacity. Our student policy successfully preserves the multi-modal topology of the expert distribution, completely avoiding the catastrophic mode collapse observed in naive single-step methods and standard distillation baselines. Extensive experiments in both simulation and dynamic real-world scenarios confirmed that our distilled policy achieves a $43\times$ speedup (operating at $\sim$125 Hz) with minimal performance degradation, successfully enabling highly reactive, closed-loop manipulation under human-induced disturbances.

Despite its efficiency, the single-step student exhibits a small but consistent performance gap relative to the flow-based teacher, reflecting the inherent trade-off of distillation. Future work will investigate improved set-level objectives and hybrid single-/multi-step refinement to further reduce this gap.

\vspace{0.1in}

{
\small
\bibliographystyle{IEEEtran}
\bibliography{main}
}


\end{document}